\documentclass{article} 

\usepackage{preprint}

\usepackage{xcolor}
\usepackage{mdwlist}

\usepackage[utf8]{inputenc} 
\usepackage[T1]{fontenc}    
\usepackage{hyperref}       
\usepackage{url}            
\usepackage{booktabs}       
\usepackage{amsfonts}       
\usepackage{nicefrac}       
\usepackage{microtype}      
\usepackage{graphicx}
\usepackage{mdwlist}        
\usepackage{import}
\usepackage{ulem}
\usepackage{url}
\usepackage{mdwlist}
\usepackage{boxedminipage}
\usepackage{listings}
\usepackage{spverbatim}
\usepackage{ntheorem}
\usepackage[separate-uncertainty=true]{siunitx}
\usepackage{diagbox}

\title{\textsc{climate-fever}: A Dataset for Verification of Real-World Climate Claims} 

\author{%
 Thomas Diggelmann  \\
 Department of Physics \\
 ETH Zurich \\
  Zurich, Switzerland \\
 \texttt{thomasdi@student.ethz.ch}
  \And
Jordan Boyd-Graber \\ 
CS, iSchool, LSC, and UMIACS\\
University of Maryland \\
College Park, MD, USA\\
\texttt{jbg@umiacs.umd.edu}
\And 
Jannis Bulian \\  
  Google Research\\
  Zurich, Switzerland \\
  \texttt{jbulian@google.com} \\
  \And
  Massimiliano Ciaramita \\
  Google Research\\
  Zurich, Switzerland \\
  \texttt{massi@google.com} \\
 \And
 Markus Leippold \\
 Department of Banking and Finance \\
 University of Zurich \\
 Zurich, Switzerland \\
  \texttt{markus.leippold@bf.uzh.ch} 
}

\newif\ifcomment\commenttrue

\ifcomment
    \newcommand{\pinaforecomment}[3]{\colorbox{#1}{\parbox{.8\linewidth}{#2: #3}}}
\else
    \newcommand{\pinaforecomment}[3]{}
\fi

\newcommand{\nothing}[1]{}

\begin{document}

\maketitle

\begin{abstract}

We introduce \textsc{climate-fever}, a new publicly available dataset for verification of climate change-related claims. By providing a dataset for the research community, we aim to facilitate and encourage work on improving algorithms for retrieving evidential support for climate-specific claims, addressing the underlying language understanding challenges, and ultimately help alleviate the impact of misinformation on climate change. We adapt the methodology of \textsc{fever} \cite{thorne2018fever}, the largest dataset of artificially designed claims, to real-life claims collected from the Internet. While during this process, we could rely on the expertise of renowned climate scientists, it turned out to be no easy task. We discuss the surprising, subtle complexity of modeling real-world climate-related claims within the \textsc{fever} framework, which we believe provides a valuable challenge for general natural language understanding. We hope that our work will mark the beginning of a new exciting long-term joint effort by the climate science and \textsc{ai} community.

\end{abstract}


\newcommand\foreign[1]{\textit{#1}}
\newtheorem{definition}{Definition}
\theoremstyle{break}
\theorembodyfont{\upshape}
\newtheorem{example}{Example}[section]

\section{Introduction}\label{sec:Introduction}

With the easy availability of information through the Internet and social media, claims of unknown veracity manipulate public perception and interpretation. 
Misinformation and disinformation are particularly pressing issues for the climate change debate. They have confused the public, led to political inaction, and stalled support for climate-change mitigation measures~\cite{anderegg2010expert, ding2011support, benegal2018correcting, van2017inoculating}. To counter the influence of potentially false claims on the formation of public opinion on climate change, researchers and experts began to manually assess claims' veracity and publish their assessments on platforms such as \url{climatefeedback.org} and \url{skepticalscience.com}.

Recently, new literature on algorithmic fact-checking has emerged, using machine learning and natural language understanding (\textsc{nlu}) to work on this problem from different angles. One influential framework that combines several of these aspects is \textsc{fever} \cite{thorne2018fever}. It consists of a well-vetted dataset of human-generated claims and evidence retrieved from Wikipedia and a shared-task for evaluating implementations of claim validators. Given that the \textsc{fever} claims are artificially constructed, they may not share the characteristics of real-world claims. Consequently, researchers~\cite{augenstein2019multifc} started to collect real-world claims from multiple fact-checking organizations along with evidence manually curated by human fact-checkers.

We believe that technology cannot and will not in the foreseeable future replace human fact-checkers. But it can help to provide relevant, reliable evidence for humans to make better decisions about the veracity of a claim. In this work, we focus on building a dataset of real-world claims specifically on climate change. We collect 1,535 claims on the Internet. For each claim, we algorithmically retrieve the top five relevant evidence candidate sentences from Wikipedia by the use of \textsc{nlu} where humans annotate each sentence as supporting, refuting, or not giving enough information to validate the claim. We call this database of 7,675 annotated claim-evidence pairs the \textsc{climate-fever} dataset.\footnote{We make this dataset publicly available at \url{http://climatefever.ai}.}.

\section{Methodology}

We adopt a pipeline approach for our evidence retrieval and claim validation system similar to the baseline system proposed by \textsc{fever} \cite{thorne2018fever} and similar to virtually all competing implementations that followed \cite{soleimani2020bert, hanselowski2018ukp, nie2019combining, zhong2019reasoning, liu2020fine}. The reason for building this system is two-fold. First, we require an algorithm to automatically retrieve evidence candidates from a large Knowledge Document Collection\footnote{We define \textsc{kdc} as any large document corpus that contains well-founded textual (prose) representations of knowledge. Examples for \textsc{kdc}'s are encyclopediae, newspaper archives, scientific publications.} (\textsc{kdc}) given a claim to build our dataset. Second, we require an end-to-end claim validation algorithm to predict entailment given a claim and evidence candidates to form a baseline, i.e., to answer the question if current claim validation approaches are up to the task of algorithmically validating real-life claims.

\begin{figure}[h]
\centering
\includegraphics[width=0.9\textwidth]{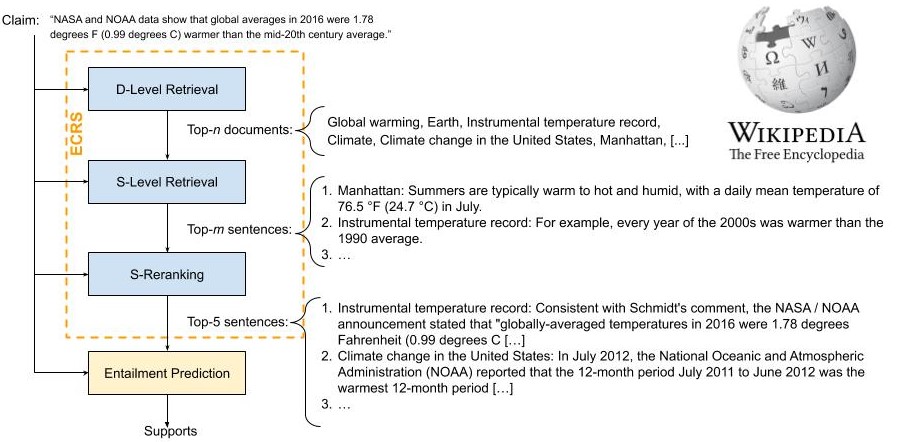}
\caption{\label{fig:pipeline} Overview of the claim validation system consisting of the Evidence Candidate Retrieval System (\textsc{ecrs}) and Entailment Prediction (\textsc{ep}) stage.}
\end{figure}

Our architecture consists of two distinct components, the Evidence Candidate Retrieval System (\textsc{ecrs}) and the Entailment Prediction (\textsc{ep}) stage (see Fig.~\ref{fig:pipeline}). For a given claim, the \textsc{ecrs} retrieves sentences as evidence candidates from the \textsc{kdc}. A pair of claim and evidence candidate sentences are fed to the \textsc{ep} to predict one of the labels \texttt{SUPPORTS}, \texttt{REFUTES} or \texttt{NOT\_ENOUGH\_INFO}, depending on whether the evidence is supporting, refuting, or not giving enough information to validate the given claim. The \textsc{fever} dataset uses a copy of the English Wikipedia containing only the introductory section of all articles as \textsc{kdc}. We also use the English Wikipedia as \textsc{kdc} but, given the complexity of real-life climate claims, allow the complete body of Wikipedia articles as a source of evidence.\footnote{This extension introduces a challenge with respect to the retrieval of evidence candidates. In the original setup, the number of sentences retrieved after the document retrieval stage is permitting for pairwise ranking against a given claim (up to 100 sentences on average in the introductory section). In our case, the number of sentences becomes intractable. We, therefore, introduce a novel technique by leveraging learned sentence-embeddings combined with a fast vector similarity index, \textsc{faiss} \cite{JDH17}, to pre-select the most relevant sentences given a claim prior to the pairwise re-ranking. This can be compared to Facebook AI's Dense Passage Retrieval (\textsc{dpr}) \cite{karpukhin2020dense} that uses a similar approach for solving Open-Domain Question Answering (\textsc{qa}) by using dense embeddings rather than BM25.
} 

\subsection{Retrieving and labeling climate claims (Task-1)}



To obtain a set of candidate climate claims from the Internet, we follow an \foreign{ad-hoc} approach and use a set of seed keywords in Google searches\footnote{For this study we did not automate this step. Instead, we manually searched for seed keywords, identified potential targets, and used Python libraries such as \texttt{requests} and \texttt{BeautifulSoup} to download, parse and clean content for subsequent processing.} to identify possible sources for such claims. We either retrieved the claims manually, or we scraped the pages automatically. We collected an equal number of \textsc{climate-fever} claims from both scientifically-informed and climate change skeptics/deniers sources. This procedure resulted in a balanced set of more than 3,000 climate claims. 

\textsc{fever} claims were written and curated by annotators, enforcing a rigorous set of requirements for writing claims. The resulting claims are self-contained, short, and syntactically simple. For example, one of the rare claims related to climate change in \textsc{fever} reads: 

\texttt{\small "The Gray wolf is threatened by global warming."} 

In contrast, a candidate claim crawled from the wild is, e.g.:

\texttt{\small ``The Intergovernmental Panel on Climate Change is misleading humanity about climate change and sea levels, and that in fact, a new solar-driven cooling period is not far off.''}

Given the complexity of real claims, we introduce the definition of a verifiable claim as follows.



\begin{definition}[Verifiable claims]
\label{def:verifiability}
A claim is potentially \emph{verifiable} if it is a) well-formed and b) subjectively investigable. 
\begin{itemize*}
    \item[a)]\label{def:wellformed} A \emph{well-formed} claim is a single English sentence, consistent, unambiguous, and complete (i.e., not much implicit knowledge is needed for comprehension by the reader). 
    \item[b)]\label{def:investigable} A claim is \emph{subjectively investigable}, if evidence could be retrieved from a knowledge document collection (\textsc{kdc}) that decreases the investigators uncertainty about the truthfulness (or falsehood) of the statement.
\end{itemize*}
\end{definition}
Equipped with our definition of verifiable claims, we asked climate scientists to label our collected claims (Task 1). For each claim, we collected up to five votes. This annotation task resulted in more than 1,535 verifiable climate claims on which there was a sufficient consensus among the annotators. We give examples of verifiable and non-verifiable claims in Appendix \ref{sec:appendix-task-1}.


\subsection{Evidence candidate retrieval}

To automatically retrieve relevant evidence candidates from Wikipedia for a given claim, we build an \textsc{ecrs} pipeline consisting of the following three steps\footnote{For BM25 in the first step we leverage Apache's Lucene (\url{https://lucene.apache.org/}), while our model implementations for steps two and three are both based on Google's \textsc{albert} model \cite{lan2019albert} by using HuggingFace's \texttt{transformers} library \cite{Wolf2019HuggingFacesTS}.}:

\emph{1. Document-level retrieval}: Given a claim as input, we retrieve the most relevant documents from the \textsc{kdc}. We apply an entity-linking approach similar to \cite{hanselowski2018ukp}. We use BM25 \cite{robertson2009probabilistic} to query an inverse document index containing all English Wikipedia articles with entity mentions extracted from the claim. We use a dependency parser to identify candidates of entity mentions, and we select the top-10 relevant articles.

\emph{2. Sentence-level retrieval}: From the selected articles, we retrieve the top-100 most relevant sentences using sentence-embeddings trained on the \textsc{fever} dataset. 
To produce task-specific sentence-embeddings, we adopt a pretrained \textsc{albert} (large-v2) model in an average-pooled Siamese-setting \cite{reimers2019sentence}. 
We apply hard positive and negative mining \cite{hermans2017defense,soleimani2020bert} to compensate for the large number of possible negative examples. 

\emph{3. Sentence re-ranking}: 
Similar to \cite{soleimani2020bert} we train a point-wise model to predict a relevance score for pairs of claim and evidence. 
We adopt a pretrained \textsc{albert} (base-v2) model with a binary classifier on the \texttt{[CLS]} token. Every evidence is classified as \emph{evidence} (1) or \emph{non-evidence} (0). During training, we use claims along with supporting and refuting evidence from the \textsc{fever} training split as examples for \emph{evidence} and we randomly sample sentences from the \textsc{fever} Wikipedia dump to provide examples for \emph{non-evidence}. During inference, we sort evidence sentences by the predicted score in descending order and select the top five sentences.


\subsection{Evidence candidate labelling (Task-2)}


In Task-2, the claims together with their top five evidence sentences as retrieved by the \textsc{ecrs} are displayed to the annotators to label it as \emph{supporting}, \emph{refuting}, or \emph{not giving enough information to validate} the claim.\footnote{Given our 1,535 climate claims and five sentences per claim, we end up with 7,675 annotations.} For each claim, we collect five individual votes per claim-evidence pair, which allows us to analyze confidence and to compute inter-annotator agreement.
During post-processing we compute a micro-verdict (for every claim-evidence pair) and a macro-label for every claim (aggregated on the five micro-verdicts). The micro-verdict is given by the majority-vote for each claim-evidence pair (or it is \texttt{NOT\_ENOUGH\_INFO} on a tie). The claim-label is by default \texttt{NOT\_ENOUGH\_INFO} unless there is supporting (\texttt{SUPPORTS}) \emph{or} refuting (\texttt{REFUTES}) evidence. If there is both supporting \emph{and} refuting evidence the claim-label is \texttt{DISPUTED}.


\subsection{Entailment prediction}

For entailment prediction, the top five candidate evidence sentences along with the corresponding Wikipedia article titles are jointly compared against the claim to predict one of the labels \texttt{SUPPORTS}, \texttt{REFUTES}, or \texttt{NOT\_ENOUGH\_INFO}. We adopt a pretrained \textsc{albert} (large-v2) with a three-way classifier on the \texttt{[CLS]} token of a concatenation of claim and evidence sentences\nothing{Pairs of (article title, evidence sentence) are concatenated, i.e. ``Title-1: Sentence-1. Title-2: Sentence-2, \ldots, Title-5: Sentence-5.'', and then encoded to form one input sequence to the \textsc{albert} tokenizer while the other input sequence being the encoded claim.}. We train the model by using the \textsc{ecrs} to retrieve the top five evidence candidates for each claim from the \textsc{fever} training split and use the gold-labels as ground-truth for optimizing the cross-entropy loss. We reach a competitive label-accuracy of 77.68\% on the \textsc{fever} dev-set using our end-to-end pipeline (\textsc{sota} label-accuracy of 79.16\% on the dev-set is reported by \cite{zhong2019reasoning}). For measuring the end-to-end performance of our claim-validation pipeline on the \textsc{climate-fever} dataset, we predict labels for all claim and evidence-set pairs and compare against the gold-labels (claim-labels) from Task-2.

\section{Discussion and future work}

To gain insight into the 1,535 climate claims, we collected several statistics about the dataset. By using a clustering technique, we identified more than 20 different topics represented by the collected claims, such as claims concerning ``climate change in the arctic'', ``sea-level rise'', and more general ones concerning ``climate\_change and global warming'' (see Appendix~\ref{sec:claim-statistics}).

The evidence labelling task (Task-2) produced a dataset of 1,535 claims with an annotated set of five evidence candidates for each claim. Each evidence sentence is labelled by at least two voters ($\num{2.4 \pm 0.7}$ voters per evidence on average). The distribution for aggregate claim-labels \texttt{SUPPORTS}, \texttt{REFUTES}, \texttt{DISPUTED}, and \texttt{NOT\_ENOUGH\_INFO} is 655 (42.67\%), 253 (16.5\%), 153 (9.97\%), and 474 (30.88\%). While \textsc{fever} only contains undisputed claims, we include claims for which both supporting and refuting evidence were found. We believe that these examples are especially useful since they appear to be a common feature of real-world claims.

Furthermore, we dealt with the limitation of \textsc{fever}-style majority-vote based aggregation for deciding on a claim label. The approach is too naïve as, in many cases, retrieved evidence covers only some facets of the claim, in which case not enough evidence is present to form a final opinion. More generally, a claim (hypothesis) can at best be refuted by contrarian evidence. Epistemologically, it is impossible to assign a final verdict. In our case, the only purpose for assigning a gold-label to each climate-claim is to measure a baseline performance of a \textsc{fever} entailment predictor on our dataset. Our dataset provides both the micro-verdict labels and the claim-labels for each claim-evidence pair.

For Task-2 we measured an average inter-annotator agreement (Krippendorff's alpha) of 0.334. This low level of agreement signifies the hardness of the task, i.e., even for human annotators, it is non-trivial to decide if an evidence candidate supports or refutes a given claim. Table~\ref{tab:disagreement-overviews} details the level of disagreement on a per-slice basis.\footnote{For Task-2 we split the 1,535 claims into 10 slices of 770 claims each (except for the last slice) such that each annotator has to label 3,850 claim-evidence pairs.} We observed that training the annotators can help raise the agreement level (e.g.\ slice 8 was labelled by two second-time annotators). Furthermore, we could also see that pairs of annotators that are experts on the topic (e.g.,  climate scientists or ML specialists) tend to have a higher average agreement (cf.\ slices 0 and 3).

\begin{table}[t]
	\centering
	\caption{\label{tab:disagreement-overviews} For evidence candidate labelling (Task-2) the 1,535 claims were split into 10 slices of 770 claims each (except for the last slice). Each slice consists of 3,850 ($770 \times 5$) claim-evidence pairs. This table lists for each slice the average number of voters, the inter-annotator agreement ($\alpha_\mathrm{Krippendorff}$), the fraction of evidence sentences with total agreement and the average entropy with respect to the select class (\texttt{SUPPORTS}, \texttt{REFUTES} or \texttt{NOT\_ENOUGH\_INFO}).}
	\begin{tabular}{rrrrrr}
		\toprule
		Slice &  Size &  Avg.\ Num.\ Voters &  $\alpha_\mathrm{Krippendorff}$ &  Total Agreement &  Avg.\ Entropy \\
		\midrule
		0 &   770 &             2.227 &  0.283 &            \textbf{0.613} &         0.266 \\
		1 &   770 &             4.019 &  0.399 &            0.423 &         0.380 \\
		2 &   770 &             2.000 &  0.522 &            \textbf{0.745} &         0.176 \\
		3 &   770 &             3.001 &  0.106 &            0.201 &         0.544 \\
		4 &   770 &             2.000 &  0.215 &            0.504 &         0.344 \\
		5 &   770 &             2.000 &  0.091 &            0.404 &         0.413 \\
		6 &   770 &             2.000 &  0.252 &            0.529 &         0.327 \\
		7 &   770 &             2.825 &  0.316 &            0.461 &         0.371 \\
		8 &   770 &             2.000 &  0.431 &            \textbf{0.635} &         0.253 \\
		9 &   745 &             2.000 &  0.229 &            0.545 &         0.315 \\
		\bottomrule
	\end{tabular}
\end{table}

The \textsc{fever}-trained entailment-predictor evaluated on the \textsc{climate-fever} dataset\footnote{To stay compatible with the \textsc{fever} methodology for evaluation we simply excluded disputed claims from our dataset.} yields the following scores (cf.\ table \ref{tab:classification-report}): label-accuracy = 38.78\%, recall = 38.78\%, precision = 56.49\%, $\mathrm{F}_1$ = 32.85\%. We computed weighted-averages for the last three metrics to compensate for the unbalanced labels. The real-life nature of the \textsc{climate-fever} dataset proves to be a challenge indicated by the low label-accuracy (38.78\% --- only slightly better than chance classification), as compared to the higher and competitive label-accuracy on the original \textsc{fever} dev-set (77.69\%). As can be seen in table \ref{confusion-matrix}, the model particularly struggles to predict \texttt{SUPPORTED} claims while it performs slightly better on predicting \texttt{REFUTED} claims. We suspect that this result can mainly be attributed to the stark qualitative differences between real-world claims of \textsc{climate-fever} and the artificial nature of \textsc{fever} claims. As argued above, real-world claims pose some unique challenges and subtleties. For instance, the claim

\texttt{\small ``The melting Greenland ice sheet is already a major contributor to rising sea level and if it was eventually lost entirely, the oceans would rise by six metres around the world, flooding many of the world’s largest cities.'' }

includes a statement that the sea level will rise six meters. An \textsc{ecrs}-retrieved evidence sentence may state a sea-level rise of $x+\epsilon$ meters. Although the numbers differ, the climate scientists labelled the evidence as supportive. There are also more demanding cases. For instance, for the claim

\texttt{\small ``An article in Science magazine illustrated that a rise in carbon dioxide did not precede a rise in temperatures, but actually lagged behind temperature rises by 200 to 1000 years.''}

the \textsc{ecrs} provides both supporting and refuting evidence and labelled as such by the annotators. Such disputed claims are absent in the \textsc{fever} dataset.\footnote{We list more examples of our \textsc{climate-fever} dataset in Appendix \ref{sec:appendix-cf}.} To develop new strategies for tackling (climate-related) disinformation, we must be able to cope with the complexity of real-life claims in general, and at the same time, account for the specific characteristics of claims related to climate change. By further extending \textsc{climate-fever} and making it publicly available, we provide a first step in the right direction and hope that our work will stimulate a new long-term joint effort by climate science and \textsc{ai} community.


\section*{Acknowledgments}

We would like to thank Thomas Hofmann (ETH Zurich) for his helpful contribution to an early draft of this paper. We also thank Christian Huggel (University of Zurich) and Reto Knutti (ETH Zurich) for all their comments and their climate science expertise. Finally, we are grateful to all climate scientists and annotators who were involved in the annotation process to build the \textsc{climate-fever} database. Research supported with Cloud \textsc{tpu}s from Google's TensorFlow Research Cloud (\textsc{tfrc}).

\bibliographystyle{IEEEtran}
\small
\bibliography{references}
\newpage




\appendix
\section*{Appendix}

\section{Claim Labelling (Task 1)}
\label{sec:appendix-task-1}

In the following, we present examples of verifiable and non-verifiable claims given definition~\ref{def:verifiability}.

\begin{example}
Observe the following three potentially verifiable claims:

    \begin{quote}
        ``NASA and NOAA data show that global averages in 2016 were 1.78 degrees F (0.99 degrees C) warmer than the mid-20th century average.''
    \end{quote}
    
    \begin{quote}
        ``The amount of carbon dioxide absorbed by the upper layer of the oceans is increasing by about 2 billion tons per year.''
    \end{quote}
    
    \begin{quote}
        ``The bushfires in Australia were caused by arsonists and a series of lightning strikes, not 'climate change'.''
    \end{quote}

The above claims are verifiable (\ref{def:verifiability}) because each claim is well-formed and there is a high probability that evidence could be retrieved from a \textsc{kdc} either supporting or refuting it.

\end{example}

\begin{example}
Observe the following claim:

\begin{quote}
``Since the beginning of the Industrial Revolution, the acidity of surface ocean waters has increased by about 30 percent.13,14 This increase is the result of humans emitting more carbon dioxide into the atmosphere and hence more being absorbed into the oceans.''
\end{quote}

This claim consists of more than a single sentence and therefore does not adhere to definition \ref{def:wellformed} and, as a consequence of this, is not verifiable. 
\end{example}

\begin{example}
Observe the following claim:

\begin{quote}
``Unprecedented climate change has caused sea level at Sydney Harbour to rise approximately 0.0 cm over the past 140 years.''
\end{quote}

This claim is not verifiable because it contains information that is inconsistent (a sea level rise of 0.0 cm) in violation of definition (\ref{def:wellformed}).
\end{example}

\begin{example}
Observe the following claim:

\begin{quote}
``CO2 emissions from all commercial operations in 2018 totaled 918 million metric tons—2.4\% of global CO2 emissions from fossil fuel use.''
\end{quote}

This statement is incomplete because for its comprehension the reader would need to know that `commercial operators' is referring to air travel.

\end{example}

\begin{example}
Observe the following claim:

\begin{quote}
``Yet nature-based solutions only receive only 2\% of all climate funding.''
\end{quote}

The above sentence is ambiguous because it is missing a subject. The collection of real claims sourced from the internet contains many examples of this type of non-verifiable claims.

\end{example}

\section{Topic distribution}
\label{sec:claim-statistics}

\begin{figure}
\centering
\def\svgwidth{0.9\textwidth}
\includegraphics[width=0.7\textwidth]{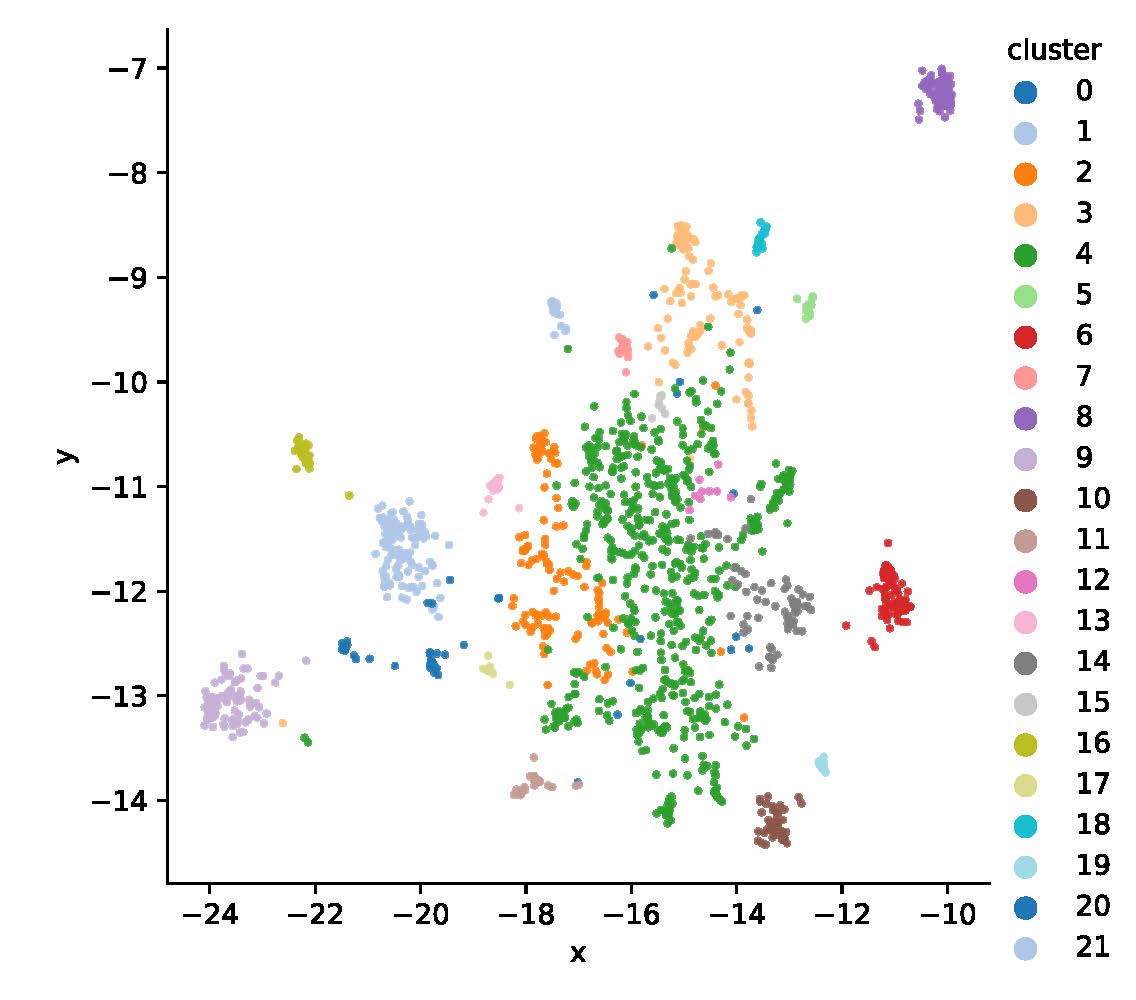} 
\caption{\label{fig:claims-scatter} Scatter plot showing a two-dimensional embedding of the 1535 climate claims using \textsc{umap} \cite{mcinnes2018umap} for dimensionality reduction. The cluster assignments were computed using the density based \textsc{dbscan} \cite{louhichi2014density} algorithm performed on 30-dimensional \textsc{umap} embeddings.}
\end{figure}

To better understand the nature of the collected climate claims from the wild we applied a clustering technique helping us to discover topics discussed in the claims. For this we pre-processed each claim by tokenizing it into its constituting tokens (words, punctuation). We then replaced each word by its lemmatized and lower-cased version. Additionally, we rejected tokens that are either stopwords or punctuation and tokens that are shorter than 3 characters. Finally, we calculated the bigrams for all words in a claim and appended these to the list of unigrams to form a total list of terms. We then built a dictionary using the pre-processed claims. We post-processed the dictionary by keeping only terms that are contained in at least 5 claims. We also rejected terms that are contained in more than 50\% of the total number of claims. After filtering, we kept the 150 most frequent terms. Table~\ref{tab:claims-most-common-tokens} lists all 150 dictionary terms sorted by document frequency, lead by the words ``global'', ``climate'' and ``warming''. We then calculated the \textsc{tf-idf} transformed document vectors for all claims using the dictionary. We applied \textsc{umap} \cite{mcinnes2018umap} to find a two-dimensional embedding of the vectors for graphical visualization as can be seen in fig.~\ref{fig:claims-scatter}. Additionally, as described in \cite{UsingUMA53:online}, we computed a 30-dimensional embedding also by applying \textsc{umap} that we used as input to the \textsc{dbscan} \cite{louhichi2014density} clustering algorithm. We identified 21+1 different clusters (21 clusters plus ambiguity cluster 0) by using this technique. Table~\ref{tab:claims-cluster-info} gives an overview about the different clusters describing cluster size and the top-5 words within the cluster measured with respect to term-frequency. It can be seen that different topics are present within the set of climate-claims represented by distinct clusters in fig.~\ref{fig:claims-scatter}, such as claims concerning ``climate change in the arctic'' (cluster 3), ``sea-level rise'' (cluster 8), and more general ones concerning ``climate\_change and global warming'' (clusters 1 and 9).

\begin{table}[t]
\centering
\caption{\label{tab:claims-most-common-tokens} Shows all 150 terms from the pruned dictionary sorted by document frequency (d.f.) for the 1,535 climate claims collected after Task-1.}
\begin{tabular}{lr|lr|lr}
\toprule
         token &  d.f. &              token &  d.f. &             token &  d.f. \\
\midrule
         global &    303 &               datum &     50 &               u.s. &     25 \\
        climate &    303 &             surface &     49 &            percent &     25 \\
        warming &    296 &               solar &     48 &                sun &     24 \\
    temperature &    222 &             weather &     45 &             united &     24 \\
 global\_warming &    189 &              energy &     45 &                bad &     24 \\
         change &    189 &                 low &     45 &           activity &     24 \\
           year &    172 &              decade &     45 &             review &     24 \\
            co2 &    158 &              recent &     43 &              sheet &     24 \\
            ice &    151 &         atmospheric &     43 &          ice\_sheet &     24 \\
          level &    143 &                long &     42 &              cycle &     24 \\
           warm &    140 &              degree &     41 &          long\_term &     24 \\
 climate\_change &    136 &              likely &     41 &                 go &     23 \\
            sea &    132 &              report &     41 &            suggest &     23 \\
           rise &    129 &                cool &     40 &               peer &     23 \\
         carbon &    128 &             predict &     40 &         scientific &     23 \\
       increase &    125 &                 new &     40 &         accelerate &     23 \\
          cause &    118 &                mean &     39 &             reduce &     23 \\
      scientist &    110 &  global\_temperature &     39 &               lead &     23 \\
          human &    105 &             sea\_ice &     38 &                air &     23 \\
       emission &     89 &              period &     37 &                age &     22 \\
          earth &     82 &                term &     37 &            ice\_age &     22 \\
      sea\_level &     82 &               event &     36 &       co2\_emission &     22 \\
        dioxide &     80 &              impact &     36 &        significant &     22 \\
 carbon\_dioxide &     80 &               occur &     34 &               fast &     22 \\
          ocean &     79 &                cold &     33 &               like &     22 \\
         record &     79 &           satellite &     33 &        peer\_review &     22 \\
     greenhouse &     72 &             extreme &     32 &           continue &     22 \\
            gas &     72 &               great &     31 &               cent &     22 \\
        century &     69 &               claim &     31 &              coral &     22 \\
           time &     67 &             average &     31 &             number &     22 \\
     atmosphere &     63 &           greenland &     31 &               20th &     22 \\
          trend &     63 &                 far &     30 &       20th\_century &     22 \\
           past &     62 &            research &     29 &               fact &     22 \\
         arctic &     61 &         measurement &     28 &            million &     22 \\
         effect &     60 &                melt &     28 &              drive &     21 \\
          model &     59 &          antarctica &     28 &               grow &     21 \\
           ipcc &     59 &              result &     28 &               reef &     21 \\
          study &     58 &                come &     28 &                ago &     21 \\
           find &     58 &               today &     27 &  climate\_scientist &     21 \\
     level\_rise &     57 &               large &     27 &              paper &     21 \\
          world &     57 &       climate\_model &     27 &              cloud &     21 \\
           high &     57 &               polar &     26 &                hot &     21 \\
            say &     55 &                 see &     26 &              major &     21 \\
        natural &     55 &                rate &     26 &              early &     21 \\
          water &     54 &                 end &     26 &          antarctic &     20 \\
           heat &     54 &               small &     26 &               half &     20 \\
       evidence &     53 &              little &     26 &            science &     20 \\
 greenhouse\_gas &     53 &              accord &     25 &            publish &     20 \\
         planet &     52 &             decline &     25 &                man &     20 \\
           show &     51 &        cause\_global &     25 &               area &     20 \\
\bottomrule
\end{tabular}
\end{table}

\begin{table}[t]
\centering
\caption{\label{tab:claims-cluster-info} Shows an overview about the different clusters found in the 1,535 climate claims. The cluster numbers correspond to the cluster numbers in Fig.~\ref{fig:claims-scatter}. The first column denotes the cluster, second column shows how many documents belong to said cluster. The last column lists the top-5 terms in the cluster as measured with respect to term-frequency.}
\begin{tabular}{rrl}
\toprule
 cluster &  size &                                              top\_terms \\
\midrule
       4 &   550 &                 co2, temperature, warm, year, increase \\
       2 &   127 &                  scientist, ipcc, climate, report, new \\
       1 &   112 &             global\_warming, warming, global, bad, year \\
       3 &   106 &                       ice, arctic, sea\_ice, polar, sea \\
       9 &   104 &          climate\_change, change, climate, human, cause \\
       6 &    78 &  carbon\_dioxide, dioxide, carbon, atmosphere, emission \\
       8 &    72 &                sea\_level, level, sea, level\_rise, rise \\
      14 &    65 &                 carbon, emission, u.s., accord, reduce \\
      10 &    51 &       greenhouse\_gas, gas, greenhouse, water, emission \\
      16 &    50 &      global\_warming, global, change, warming, increase \\
      20 &    32 &   global\_warming, cause\_global, warming, global, cause \\
      21 &    26 &                     term, long\_term, long, trend, cool \\
      18 &    24 &           ice\_sheet, sheet, ice, greenland, antarctica \\
      11 &    21 &                   extreme, event, weather, bad, change \\
      13 &    21 &                 review, peer\_review, peer, paper, ipcc \\
       7 &    19 &                        reef, coral, great, world, year \\
      19 &    19 &        20th, 20th\_century, century, early, temperature \\
       5 &    19 &                        age, ice\_age, ice, little, come \\
       0 &    14 &                       warming, time, event, trend, ice \\
      12 &     9 &                 like, temperature, degree, earth, warm \\
      17 &     9 &                    cloud, sun, predict, likely, global \\
      15 &     7 &               large, decline, area, temperature, water \\
\bottomrule
\end{tabular}
\end{table}

\section{Baseline evaluation}

To evaluate a baseline system on \textsc{climate-fever}, we trained a claim validation system on the \textsc{fever} task reaching a competitive label-accuracy of 77.58\% on the \textsc{fever} dev-set. The final entailment prediction task was formulated as a three-way classifier predicting label \texttt{SUPPORTS}, \texttt{REFUTES} or \texttt{NOT\_ENOUGH\_INFO} based on the claim and a concatenation of all evidence sentences (were we prepended the Wikipedia article title to each evidence sentence to resolve missing co-references). Tables \ref{tab:classification-report} and \ref{confusion-matrix} compare the performance of the claim validator evaluated on \textsc{climate-fever} and on the \textsc{fever} dev-set.

\begin{table}[t]
\centering
\caption{\label{tab:classification-report} Classification report with respect to the performance of the baseline claim validation pipeline evaluated on \textsc{climate-fever}. These values are contrasted (in parentheses) with the reported values from the evaluation on the original \textsc{fever} dev-set. All values are reported as percentages.}
\begin{tabular}{lrl|rl|rl|rl}
\toprule
{} & \multicolumn{2}{c}{precision} & \multicolumn{2}{c}{recall} &  \multicolumn{2}{c}{f1-score} &  \multicolumn{2}{c}{support} \\
\midrule
SUPPORTS        &       35.04 &     (81.34) &    75.11 &  (86.32) &       47.79 &    (83.76) &        474 &    (6666) \\
REFUTES         &       39.93 &      (79.2) &    43.87 &  (77.93) &       41.81 &    (78.56) &        253 &    (6666) \\
NOT ENOUGH INFO &       78.41 &     (72.08) &    10.53 &  (68.83) &       18.57 &    (70.42) &        655 &    (6666) \\
\midrule
weighted avg    &       56.49 &     (77.54) &    38.78 &  (77.69) &       32.85 &    (77.58) &       1382 &   (19998) \\
accuracy        &             &             &          &          &       38.78 &    (77.69) &       1382 &   (19998) \\
\bottomrule
\end{tabular}
\end{table}

\begin{table}[t]
\centering
\caption{\label{confusion-matrix} Normalized confusion matrix comparing classification performance of baseline claim validator evaluated on \textsc{climate-fever} and original \textsc{fever} dev-set (in parentheses). Rows correspond to true labels, columns correspond to predicted labels. All values are reported as percentages.}
\begin{tabular}{lrr|rr|rr}
\toprule
{} &  \multicolumn{2}{c}{SUPPORTS} & \multicolumn{2}{c}{REFUTES} &  \multicolumn{2}{c}{NOT ENOUGH INFO} \\
\midrule
SUPPORTS        & 10.5 & (86.3) & 9.2 & (4.2) & 80.3 & (9.5) \\
REFUTES         & 3.2 & (4.9) & 43.9 & (77.9) & 53.0 & (17.2) \\
NOT ENOUGH INFO &      2.3 & (14.9) & 22.6 & (16.2) & 75.1 & (68.8) \\
\bottomrule
\end{tabular}
\end{table}

\section{\textsc{climate-fever} dataset}
\label{sec:appendix-cf}

In this section we give an overview about the collected and labelled claims from the \textsc{climate-fever} dataset. We show examples of supported and refuted claims and also give examples for disputed statements and claims that are not-verifiable (should have been rejected during Task-1). We believe that one draw-back of the original \textsc{fever} dataset is the lack of examples with contradictory evidence which naturally seem to arise when dealing with real-life claims. We also report the average entropy for each claim as calculated by interpreting the relative frequencies of label votes as label membership probabilities and calculating the mean entropy with respect to the individual entropies calculated for the evidence sentences. Entropy then acts as surrogate for measuring the inter-annotator agreement: high entropy means disagreement, low entropy means agreement. The entropy is naturally zero for claims were we so far only collected a single vote per evidence.

We note that the claim-label is \texttt{SUPPORTS}, if at least one micro-verdict is \texttt{SUPPORTS} and all others are \texttt{NOT\_ENOUGH\_INFO}; it is \texttt{REFUTES}, if at least one micro-verdict is \texttt{REFUTES} and all others are \texttt{NOT\_ENOUGH\_INFO}; it is \texttt{NOT\_ENOUGH\_INFO}, if all micro-verdicts are \texttt{NOT\_ENOUGH\_INFO}; otherwise the claim-label is \texttt{DISPUTED}.

\subsection{Supported claims}

The following claims were all supported by evidence sentences retrieved by the \textsc{ecrs} as labelled by the annotators.

\begin{minipage}{\textwidth}
\begin{example}
\label{ex:claim-sup-1}

Here, we believe that the high inter-annotator disagreement is due to the incoherent formulation of the claim (``more than 100 per cent \ldots''). However, it is still clear what the statement intends to say which is why it correctly was labelled as verifiable during Task-1.

\hfill

\begin{boxedminipage}{\textwidth}
\begin{description*}
\item[Votes]: 4
\item[Entropy:] 1.04
\item[Claim:] more than 100 per cent of the warming over the past century is due to human actions
\item[Evidence:]\hfill{
\begin{description*}
\item[Supports:] The view that human activities are likely responsible for most of the observed increase in global mean temperature (""global warming"") since the mid-20th century is an accurate reflection of current scientific thinking. [\texttt{wiki/Kyoto\_Protocol}]
\item[Not\_Enough\_Info:] Human-caused increases in greenhouse gases are responsible for most of the observed global average surface warming of roughly 0.8$^{\circ}$C (1.5$^{\circ}$F) over the past 140 years. [\texttt{wiki/Scientific\_consensus\_on\_climate\_change}]
\item[Supports:] The dominant cause of the warming since the 1950s is human activities. [\texttt{wiki/Scientific\_consensus\_on\_climate\_change}]
\item[Supports:] The global warming observed over the past 50 years is due primarily to human-induced emissions of heat-trapping gases. [\texttt{wiki/Scientific\_consensus\_on\_climate\_change}]
\item[Supports:] Human activities, primarily the burning of fossil fuels (coal, oil, and natural gas), and secondarily the clearing of land, have increased the concentration of carbon dioxide, methane, and other heat-trapping (""greenhouse"") gases in the atmosphere...There is international scientific consensus that most of the warming observed over the last 50 years is attributable to human activities. [\texttt{wiki/Scientific\_consensus\_on\_climate\_change}]
\end{description*}
}\item[Verdict:] Supports
\end{description*}
\end{boxedminipage}

\end{example}
\end{minipage}

\begin{minipage}{\textwidth}
\begin{example}
\hfill

\begin{boxedminipage}{\textwidth}
\begin{description*}
\item[Votes]: 4
\item[Entropy:] 0.0
\item[Claim:] A paper by Ross McKitrick, an economics professor at the  University of Guelph, and Patrick Michaels, an environmental studies  professor at the University of Virginia, concludes that half of the  global warming trend from 1980 to 2002 is caused by Urban Heat Island.
\item[Evidence:]\hfill{
\begin{description*}
\item[Not\_Enough\_Info:] McIntyre agreed, and made contact with University of Guelph economics professor Ross McKitrick, a senior fellow of the Fraser Institute which opposed the Kyoto treaty, and co-author of Taken By Storm: The Troubled Science, Policy and Politics of Global Warming. [\texttt{wiki/Hockey\_stick\_controversy}]
\item[Not\_Enough\_Info:] A 2002 article published in the journal Climate Research by Michaels and three other scholars has predicted ""a warming range of 1.3–3.0$^{\circ}$C, with a central value of 1.9$^{\circ}$C"" over the 1990 to 2100 period, although he remarked that the ""temperature range and central values determined in our study may be too great."" [\texttt{wiki/Patrick\_Michaels}]
\item[Not\_Enough\_Info:] Until 2007 he was research professor of environmental sciences at the University of Virginia, where he had worked from 1980. [\texttt{wiki/Patrick\_Michaels}]
\item[Not\_Enough\_Info:] McKitrick has authored works about environmental economics and climate change issues, including co-authoring the book Taken by Storm: The Troubled Science, Policy and Politics of Global Warming, published in 2002. [\texttt{wiki/Ross\_McKitrick}]
\item[Supports:] For example, Ross McKitrick and Patrick J. Michaels conducted a statistical study of surface-temperature data regressed against socioeconomic indicators, and concluded that about half of the observed warming trend (for 1979–2002) could be accounted for by the residual UHI effects in the corrected temperature data set they studied—which had already been processed to remove the (modeled) UHI contribution. [\texttt{wiki/Urban\_heat\_island}]
\end{description*}
}\item[Verdict:] Supports
\end{description*}
\end{boxedminipage}

\end{example}
\end{minipage}

\subsection{Refuted claims}

The following claims were all refuted by evidence sentences retrieved by the \textsc{ecrs} as labelled by the annotators.

\begin{minipage}{\textwidth}
\begin{example}
\label{ex:claim-ref-1}
In this example, there is a high inter-annotator disagreement due to the second evidence sentence. We believe that the reason for the disagreement is because some of the annotators were aware of the popular original statement that specifically refers to the consensus among climate scientists (and not the US population which is the subject of the second evidence sentence). Additionally, the percentages given differ to quite a large extent which might also have contributed to the disagreement. 

\hfill

\begin{boxedminipage}{\textwidth}
\begin{description*}
\item[Votes]: 4
\item[Entropy:] 0.85
\item[Claim:] 97\% consensus on human-caused global warming has been disproven.
\item[Evidence:]\hfill{
\begin{description*}
\item[Not\_Enough\_Info:] In the scientific literature, there is an overwhelming consensus that global surface temperatures have increased in recent decades and that the trend is caused mainly by human-induced emissions of greenhouse gases. [\texttt{wiki/Global\_warming}]
\item[Refutes:] In a 2019 CBS poll, 64\% of the US population said that climate change is a ""crisis"" or a ""serious problem"", with 44\% saying human activity was a significant contributor. [\texttt{wiki/Global\_warming}]
\item[Refutes:] Of these, 97\% agree, explicitly or implicitly, that global warming is happening and is human-caused. [\texttt{wiki/Scientific\_consensus\_on\_climate\_change}]
\item[Not\_Enough\_Info:] It is extremely likely (95–100\% probability) that human influence was the dominant cause of global warming between 1951–2010. [\texttt{wiki/Scientific\_consensus\_on\_climate\_change}]
\item[Refutes:] 97\% of the scientists surveyed agreed that global temperatures had increased during the past 100 years; 84\% said they personally believed human-induced warming was occurring, and 74\% agreed that ""currently available scientific evidence"" substantiated its occurrence. [\texttt{wiki/Scientific\_consensus\_on\_climate\_change}]
\end{description*}
}\item[Verdict:] Refutes
\end{description*}
\end{boxedminipage}

\end{example}
\end{minipage}

\begin{minipage}{\textwidth}
\begin{example}
\hfill

\begin{boxedminipage}{\textwidth}
\begin{description*}
\item[Votes]: 4
\item[Entropy:] 0.23
\item[Claim:] Extreme weather isn't caused by global warming
\item[Evidence:]\hfill{
\begin{description*}
\item[Refutes:] Extreme Weather Prompts Unprecedented Global Warming Alert. [\texttt{wiki/Extreme\_weather}]
\item[Refutes:] Scientists attribute extreme weather to man-made climate change. [\texttt{wiki/Extreme\_weather}]
\item[Refutes:] Researchers have for the first time attributed recent floods, droughts and heat waves, to human-induced climate change. [\texttt{wiki/Extreme\_weather}]
\item[Refutes:] Climate change is more accurate scientifically to describe the various effects of greenhouse gases on the world because it includes extreme weather, storms and changes in rainfall patterns, ocean acidification and sea level."". [\texttt{wiki/Global\_warming}]
\item[Refutes:] The effects of global warming include rising sea levels, regional changes in precipitation, more frequent extreme weather events such as heat waves, and expansion of deserts. [\texttt{wiki/Global\_warming}]
\end{description*}
}\item[Verdict:] Refutes
\end{description*}
\end{boxedminipage}

\end{example}
\end{minipage}

\subsection{Disputed claims}

For the following claims contradictory evidence was found by the \textsc{ecrs} as labelled by the annotators. Such examples are especially interesting since the original \textsc{fever} dataset does not contain such examples. We believe that in the future it is important to extend the dataset with disputed examples, such that an end-to-end pipeline can also predict cases like these.

\begin{minipage}{\textwidth}
\begin{example}
\hfill

\begin{boxedminipage}{\textwidth}
\begin{description*}
\item[Votes]: 4
\item[Entropy:] 0.23
\item[Claim:] ""An article in Science magazine illustrated that a rise in carbon dioxide did not precede a rise in temperatures, but actually lagged behind temperature rises by 200 to 1000 years.
\item[Evidence:]\hfill{
\begin{description*}
\item[Not\_Enough\_Info:] In 2019 a paper published in the journal Science found the oceans are heating 40\% faster than the IPCC predicted just five years before. [\texttt{wiki/Effects\_of\_global\_warming}]
\item[Supports:] Studies of the Vostok ice core show that at the ""beginning of the deglaciations, the CO 2 increase either was in phase or lagged by less than ~1000 years with respect to the Antarctic temperature, whereas it clearly lagged behind the temperature at the onset of the glaciations"". [\texttt{wiki/Global\_warming\_controversy}]
\item[Refutes:] Recent warming is followed by carbon dioxide levels with only a 5 months delay. [\texttt{wiki/Global\_warming\_controversy}]
\item[Not\_Enough\_Info:] Temperatures rose by 0.0$^{\circ}$C–0.2$^{\circ}$C from 1720–1800 to 1850–1900 (Hawkins et al., 2017). [\texttt{wiki/Global\_warming}]
\item[Not\_Enough\_Info:] Carbon dioxide concentrations were relatively stable for the past 10,000 years but then began to increase rapidly about 150 years ago…as a result of fossil fuel consumption and land use change. [\texttt{wiki/Scientific\_consensus\_on\_climate\_change}]
\end{description*}
}\item[Verdict:] Disputed
\end{description*}
\end{boxedminipage}

\end{example}
\end{minipage}

\begin{minipage}{\textwidth}
\begin{example}
\hfill

\begin{boxedminipage}{\textwidth}
\begin{description*}
\item[Votes]: 4
\item[Entropy:] 0.66
\item[Claim:] Droughts and floods have not changed since we've been using fossil fuels
\item[Evidence:]\hfill{
\begin{description*}
\item[Not\_Enough\_Info:] According to the WWF, the combination of climate change and deforestation increases the drying effect of dead trees that fuels forest fires. [\texttt{wiki/Drought}]
\item[Supports:] However, other research suggests that there has been little change in drought over the past 60 years. [\texttt{wiki/Effects\_of\_global\_warming}]
\item[Refutes:] Due to deforestation the rainforest is losing this ability, exacerbated by climate change which brings more frequent droughts to the area. [\texttt{wiki/Effects\_of\_global\_warming}]
\item[Refutes:] There may have been changes in other climate extremes (e.g., floods, droughts and tropical cyclones) but these changes are more difficult to identify. [\texttt{wiki/Effects\_of\_global\_warming}]
\item[Refutes:] The increased demands are contributing to increased environmental degradation and to global warming, with resultant intensification of tropical cyclones, floods, droughts, forest fires, and incidence of hyperthermia deaths. [\texttt{wiki/History\_of\_the\_world}]
\end{description*}
}\item[Verdict:] Disputed
\end{description*}
\end{boxedminipage}

\end{example}
\end{minipage}

\subsection{Subtle cases of claims and evidences}

Here we list some interesting claims that showcase the challenges with real-life statements.

\begin{minipage}{\textwidth}
\begin{example}
Sometimes the decision if a sentence is supporting or refuting a claim is subtle as in this case. The quantification ``six metres'' in the statement is not directly echoed in the evidence sentences (all evidence candidates mention 7 metres). However, the general claim is still supported.

\hfill

\begin{boxedminipage}{\textwidth}
\begin{description*}
\item[Votes]: 1
\item[Entropy:] 0.0
\item[Claim:] The melting Greenland ice sheet is already a major contributor to rising sea level and if it was eventually lost entirely, the oceans would rise by six metres around the world, flooding many of the world’s largest cities.
\item[Evidence:]\hfill{
\begin{description*}
\item[Supports:] If the entire 2,850,000 km3 (684,000 cu mi) of ice were to melt, global sea levels would rise 7.2 m (24 ft). [\texttt{wiki/Greenland\_ice\_sheet}]
\item[Supports:] Ice sheet models project that such a warming would initiate the long-term melting of the ice sheet, leading to a complete melting of the ice sheet (over centuries), resulting in a global sea level rise of about 7 metres (23 ft). [\texttt{wiki/Greenland\_ice\_sheet}]
\item[Supports:] If the entire 2,850,000 cubic kilometres (684,000 cu mi) of ice were to melt, it would lead to a global sea level rise of 7.2 m (24 ft). [\texttt{wiki/Greenland\_ice\_sheet}]
\item[Supports:] If the Greenland ice sheet were to melt away completely, the world's sea level would rise by more than 7 m (23 ft). [\texttt{wiki/Greenland}]
\item[Supports:] The Greenland ice sheet occupies about 82\% of the surface of Greenland, and if melted would cause sea levels to rise by 7.2 metres. [\texttt{wiki/Ice\_sheet}]
\end{description*}
}\item[Verdict:] Supports
\end{description*}
\end{boxedminipage}

\end{example}
\end{minipage}

\begin{minipage}{\textwidth}
\begin{example}

This is another example of the subtleties of interpreting evidence sentences with respect to the given claim. Here the claim is clearly referring to polar ice, but a majority of the evidence sentences are talking about glacial retreat. However, the later information is only discovered if one is careful enough to realize the Wikipedia article title from which the evidence sentences are extracted (\texttt{wiki/Retreat\_of\_glaciers\_since\_1850}).

\hfill

\begin{boxedminipage}{\textwidth}
\begin{description*}
\item[Votes]: 1
\item[Entropy:] 0.0
\item[Claim:] Beginning in 2005, however, polar ice modestly receded for several years.
\item[Evidence:]\hfill{
\begin{description*}
\item[Refutes:] Polar Discovery "Continued Sea Ice Decline in 2005". [\texttt{wiki/Arctic\_Ocean}]
\item[Not\_Enough\_Info:] Ice cover decreased to 297 km2 (115 sq mi) by 1987–1988 and to 245 km2 (95 sq mi) by 2005, 50\% of the 1850 area. [\texttt{wiki/Retreat\_of\_glaciers\_since\_1850}]
\item[Not\_Enough\_Info:] The net loss in volume and hence sea level contribution of the Greenland Ice Sheet (GIS) has doubled in recent years from 90 km3 (22 cu mi) per year in 1996 to 220 km3 (53 cu mi) per year in 2005. [\texttt{wiki/Retreat\_of\_glaciers\_since\_1850}]
\item[Not\_Enough\_Info:] The Trift Glacier had the greatest recorded retreat, losing 350 m (1,150 ft) of its length between the years 2003 and 2005. [\texttt{wiki/Retreat\_of\_glaciers\_since\_1850}]
\item[Not\_Enough\_Info:] This long-term average was markedly surpassed in recent years with the glacier receding 30 m (98 ft) per year during the period between 1999–2005. [\texttt{wiki/Retreat\_of\_glaciers\_since\_1850}]
\end{description*}
}\item[Verdict:] Refutes
\end{description*}
\end{boxedminipage}

\end{example}
\end{minipage}

\end{document}